%% file: main.tex
\documentclass[10pt,twocolumn,letterpaper]{article}

\usepackage[pagenumbers]{cvpr} % To force page numbers, e.g. for an arXiv version

\usepackage{makecell}
\usepackage{colortbl}

\input{preamble}
\definecolor{cvprblue}{rgb}{0.21,0.49,0.74}
\usepackage[pagebackref,breaklinks,colorlinks,allcolors=cvprblue]{hyperref}

\def\paperID{5039} % *** Enter the Paper ID here
\def\confName{CVPR}
\def\confYear{2026}

\title{Toward Real-world Infrared Image Super-Resolution: A Unified Autoregressive Framework and Benchmark Dataset}

\author{
  Yang Zou\textsuperscript{\rm 1}$^{\dagger}$, \quad Jun Ma\textsuperscript{\rm 2}$^{\dagger}$, \quad Zhidong Jiao\textsuperscript{\rm 2}, \quad Xingyuan Li\textsuperscript{\rm 3}, \quad Zhiying Jiang\textsuperscript{\rm 4}, \quad Jinyuan Liu\textsuperscript{\rm 2}\thanks{Corresponding author.}\\
  {\small\textsuperscript{1} School of Computer Science, Northwestern Polytechnical University}\\
  {\small\textsuperscript{2} School of Software Technology \& DUT-RU International School of ISE, Dalian University of Technology} \\
  {\small\textsuperscript{3} School of Computer Science, Zhejiang University}\\
  {\small\textsuperscript{4} College of Information Science and Technology, Dalian Maritime University}\\
  {\tt\small archerv2@mail.nwpu.edu.cn} \hspace{0.1cm}
  {\tt\small atlantis918@hotmail.com} \hspace{0.1cm}
}

\makeatletter
\def\blfootnote{\gdef\@thefnmark{}\@footnotetext}
\makeatother

\begin{document}
% \maketitle
\twocolumn[{%
\renewcommand\twocolumn[1][]{#1}%
\maketitle
\begin{center}
    \centering
    \captionsetup{type=figure}
    \vspace{-0.25in}
    \includegraphics[width=1\textwidth]{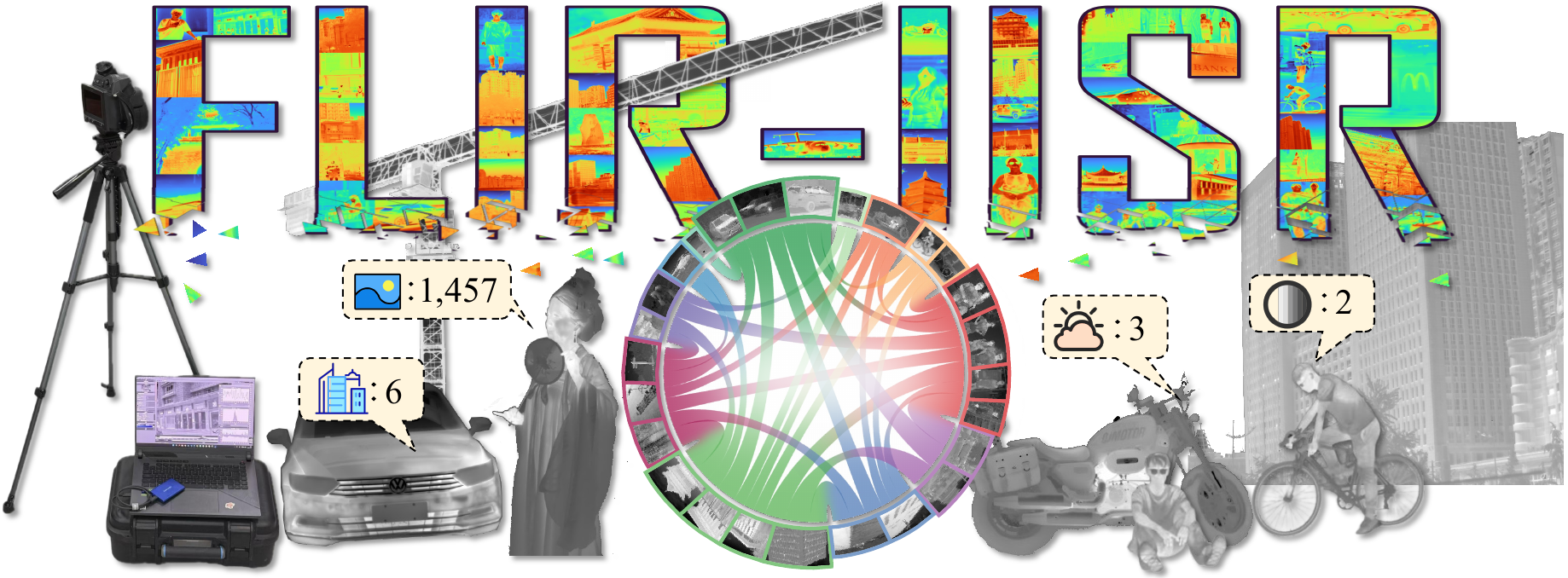}
    \vspace{-0.2in}
    \captionof{figure}{Overview of the constructed \textbf{FLIR-IISR} dataset. To bridge the gap between synthetic and real-world infrared image super-resolution (IISR), we construct the FLIR-IISR dataset using a FLIR T1050sc camera at a \textbf{1024×768} resolution, which contains \textbf{1,457} paired LR–HR images captured across \textbf{6} cities, \textbf{3} seasons, and \textbf{2} real blur patterns of optical and motion blur across \textbf{12} scene categories. 
}
    \label{fig:teaser}
\end{center}%
}]

\input{sec/0_abstract}    
\input{sec/1_intro}

\input{sec/2_related_works}
\input{sec/3_method}

\input{sec/4_experiments}

\input{sec/5_conclusion}

\section*{Acknowledgments}
This work was partially supported by the China Postdoctoral Science Foundation (2023M730741) and the National Natural Science Foundation of China (No.62302078, No.62372080).

{
    \small
    \bibliographystyle{ieeenat_fullname}
    \bibliography{main}
}


\end{document}

%% file: sec/0_abstract.tex
\begin{abstract}

\blfootnote{$^{\dagger}$ Equal contribution. $^*$Corresponding author.}

Infrared image super-resolution (IISR) under real-world conditions is a practically significant yet rarely addressed task. Pioneering works are often trained and evaluated on simulated datasets or neglect the intrinsic differences between infrared and visible imaging. In practice, however, real infrared images are affected by coupled optical and sensing degradations that jointly deteriorate both structural sharpness and thermal fidelity. To address these challenges, we propose \textbf{Real-IISR}, a unified autoregressive framework for real-world IISR that progressively reconstructs fine-grained thermal structures and clear backgrounds in a scale-by-scale manner via thermal-structural guided visual autoregression. Specifically, a Thermal-Structural Guidance module encodes thermal priors to mitigate the mismatch between thermal radiation and structural edges. Since non-uniform degradations typically induce quantization bias, Real-IISR adopts a Condition-Adaptive Codebook that dynamically modulates discrete representations based on degradation-aware thermal priors. Also, a Thermal Order Consistency Loss enforces a monotonic relation between temperature and pixel intensity, ensuring relative brightness order rather than absolute values to maintain physical consistency under spatial misalignment and thermal drift. We build \textbf{FLIR-IISR}, a real-world IISR dataset with paired LR-HR infrared images acquired via automated focus variation and motion-induced blur. Extensive experiments demonstrate the promising performance of Real-IISR, providing a unified foundation for real-world IISR and benchmarking. The dataset and code are available at:
\url{https://github.com/JZD151/Real-IISR}.
\end{abstract}

%% file: sec/1_intro.tex
\section{Introduction}
\label{sec:intro}
Infrared image super-resolution (IISR) is essential for various perception tasks like object detection, target tracking, and autonomous driving under low-light or adverse conditions~\cite{li2024contourlet, li2025difiisr, zou2026contourlet, yang2025instruction, li2025mulfs}. Recent advances in real-world image super-resolution~\cite{ji2020real, wang2024sinsr, qu2025visual, wu2024seesr} have substantially improved the reconstruction quality of visible images. However, extending such progress to the infrared domain remains highly non-trivial. The longer wavelengths and weaker atmospheric scattering in infrared sensing lead to spatially varying blur, unstable thermal boundaries, and temperature-dependent radiometric drift, which together form complex, coupled degradations~\cite{liu2025toward, liu2025dcevo, liu2024promptfusion, wang2025efficient}. These challenges make real-world IISR a unique and fundamentally difficult problem.

Previous studies on real-world image super-resolution (ISR) have largely addressed complex degradations. RealSR~\cite{cai2019toward} pioneered paired data collection via focal-length variation, while BSRGAN~\cite{zhang2021designing} and Real-ESRGAN~\cite{wang2021real} introduced effective degradation pipelines for blind ISR. Building on these advances, diffusion-based~\cite{wang2024sinsr,wang2024exploiting,wu2024seesr,yang2024pixel,lin2024diffbir, shi2024vdmufusion, fang2024real} and visual autoregressive~\cite{qu2025visual,kong2025nsarm} models further improved perceptual fidelity. However, the stochastic sampling of diffusion models and the absence of infrared degradation priors limit their applicability to real-world IISR~\cite{wu2024seesr,wang2024exploiting,qu2025visual, fang2025integrating, ma2024follow}.

On the other hand, recent IISR methods exploit sensing-specific properties to compensate for weak high-frequency details. ChasNet~\cite{prajapati2021channel} enhances informative channels via channel-split convolutions, while CoRPLE~\cite{li2024contourlet} and CRG~\cite{zou2026contourlet} introduce contourlet-domain residual modeling with prompt guidance. DifIISR~\cite{li2025difiisr} further integrates gradient-based alignment and perceptual priors within diffusion. Despite their progress, these approaches rely on simplified degradations, limiting their robustness to complex real-world infrared degradations.

While prior studies have advanced real-world and infrared super-resolution (SR), two fundamental challenges remain for real-world IISR. \textbf{(1) Lack of real infrared degradation datasets.} Existing IISR methods are typically trained on downsampled Infrared and Visible Image Fusion (IVIF) datasets, which fail to capture the coupled optical–sensor degradations of real infrared imaging, leading to poor generalization. \textbf{(2) Absence of infrared-aware degradation modeling.} Diffusion-based SR networks rely on fixed degradation priors, overlooking spatially heterogeneous blur and noise~\cite{wu2024seesr,wang2024exploiting,qu2025visual, ma2025follow, li2023text}. Meanwhile, visual autoregressive models~\cite{qu2025visual, kong2025nsarm} are confined to visible images, lacking infrared-specific constraints where thermal intensity often misaligns with structural edges, causing boundary distortion and thermal drift.

To this end, we construct a real-world IISR dataset, \textbf{FLIR-IISR}, inspired by RealSR~\cite{cai2019toward}. As shown in \cref{fig:teaser}, it contains 1,457 real captured LR–HR pairs acquired with a FLIR T1050sc camera at a 1024×768 resolution, spanning 12 scenes across 6 cities, and 3 seasons. LR images were obtained by automated focus variation and real object motion to produce realistic defocus and motion blur degradations, providing a new benchmark for real-world IISR.

We also propose \textbf{Real-IISR}, a unified autoregressive framework for real-world IISR. Specifically, Real-IISR proposes a Thermal-Structural Guidance (TSG) module that explicitly encodes thermal intensity and edge-aware structural cues to bridge the inherent mismatch between heat distributions and object boundaries, enabling structure-consistent and thermally stable reconstruction. To correct quantization bias and maintain thermal fidelity under spatially nonuniform blur and noise, Real-IISR introduces a Condition-Adaptive Codebook (CAC) that dynamically modulates discrete embeddings according to infrared degradation priors. Given the monotonic property of infrared imaging, where higher temperatures correspond to higher pixel intensities~\cite{hudson1969infrared, rieke2008absolute}, Real-IISR further adopts a Thermal Order Consistency Loss that enforces relative order consistency between patch pairs and remains robust to LR–HR misalignment in real-world infrared data. Our contributions are summarized as follows:

\begin{itemize}
    \item We construct \textbf{FLIR-IISR}, a real-world IISR dataset comprising 1,457 LR–HR pairs with real-world degradations, thereby providing a new benchmark for real-world IISR.
    \item We propose \textbf{Real-IISR}, a unified autoregressive framework guided by thermal priors that adaptively handles heterogeneous infrared degradations.
    \item Extensive experiments on both the proposed FLIR-IISR dataset and simulated dataset demonstrate the impressive performance of our method.
\end{itemize}

%% file: sec/2_related_works.tex
\begin{figure*}[!htp]
    \centering
    \includegraphics[width=1\linewidth]{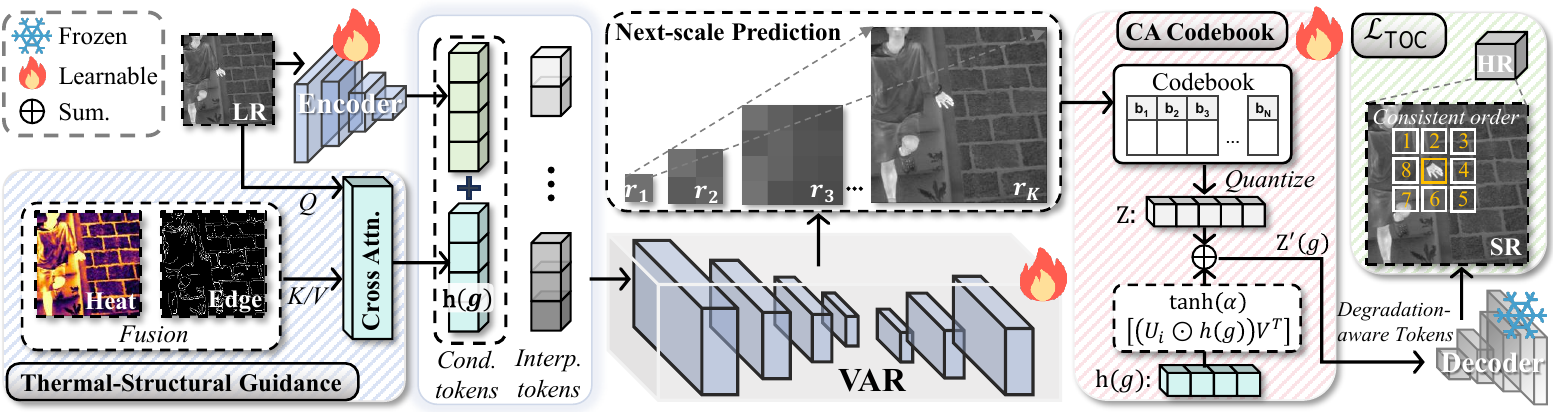}
    \caption{Overview of Real-IISR. The Thermal-Structural Guidance (TSG) module fuses thermal priors for degradation-aware encoding. The VAR backbone performs scale-by-scale generation via next-scale prediction, while the Condition-Adaptive Codebook (CAC) dynamically adjusts quantized embeddings based on degradation-aware priors for thermal fidelity. Finally, the Thermal Order Consistency Loss \(\mathcal{L}_{\text{TOC}}\) preserves physically consistent thermal ordering.}
    \label{fig:pipeline}
\end{figure*}

\section{Related work}
\label{sec:related_works}
\subsection{Image Super-Resolusion}
Early methods~\cite{liu2022video,zhang2018image,wang2020deep} assume simple degradations, which limit their generalization to real-world conditions. RealSR~\cite{cai2019toward} mitigates this gap by capturing paired LR–HR samples under varying focal lengths, while later works~\cite{zhang2018learning,xu2020unified} improve realism yet remain sensitive to spatially variant blur and compound noise. Generative frameworks, including GAN-based~\cite{wang2021real,zhang2021designing}, diffusion-based~\cite{song2020score,wang2024sinsr}, and autoregressive models~\cite{qu2025visual,kong2025nsarm,sanchez2025multi}, further enhance perceptual fidelity via learned priors.

Extending such approaches to infrared imaging remains challenging due to wavelength-dependent blur, nonlinear radiometric responses, and unstable thermal boundaries~\cite{qu2024frequency,huang2025infrared,meng2025modeling,zhou2025adversarially,zou2026hatir}. IISR methods, such as ChasNet~\cite{prajapati2021channel}, CoRPLE~\cite{li2024contourlet}, CRG~\cite{zou2026contourlet}, and DifIISR~\cite{li2025difiisr}, largely address these issues but often rely on synthetic degradations, hindering their robustness in real-world conditions.

\subsection{Visual Autoregressive Models}
Visual Autoregressive (VAR) models model conditional dependencies for token-wise generation with global structural control.
Early patch-based methods~\cite{van2017neural,oussidi2018deep} suffered from limited receptive fields and poor semantic consistency. Subsequent hierarchical designs such as VQ-VAE-2~\cite{razavi2019generating} and VQGAN~\cite{esser2021taming} introduced multi-scale codebooks and adversarial quantization, improving perceptual fidelity and enabling scale-by-scale generation. The VAR framework~\cite{tian2024visual} further formalized “next-scale prediction,” achieving efficient coarse-to-fine synthesis.
Applied to SR, VARSR~\cite{qu2025visual} integrates positive-negative pairs for stable reconstruction, yet its unified sampling remains susceptible to local blur and texture breaks under complex degradations. 

%% file: sec/3_method.tex
\section{Method}
\noindent \textbf{Overview.} Real-world IISR is inherently challenging due to complex optical–sensor degradations and the weak correlation between thermal intensity and structural edges. To address these issues, we propose a unified autoregressive framework for high-fidelity and physically consistent IISR, as illustrated in \cref{fig:pipeline}. The framework comprises three key components: (1) Thermal-Structural Guidance (TSG), which explicitly encodes heat source semantics and structural edges to align thermal distributions with spatial boundaries; (2) Condition-Adaptive Codebook (CAC), which dynamically modulates codebook embeddings with degradation-aware priors to enhance texture realism and robustness under diverse degradations; and (3) Thermal Order Consistency Loss \(\mathcal{L}_{\text{TOC}}\), which preserves the monotonic thermal–intensity relationship between SR and HR, mitigating thermal drift and boundary distortion. Together, these components enable our model to produce geometrically consistent, texture-rich, and thermally reliable infrared reconstructions under real-world degradations.

\subsection{Thermal-Structural Guidance}
Infrared imaging inherently suffers from the weak correspondence between thermal intensity and structural boundaries. For instance, although a car engine acts as a strong heat source, its thermal radiation region often deviates from the actual contour of the vehicle. Training a generative model directly on such images can cause it to overfit thermal peaks while neglecting real edges, leading to structural distortion and thermal drift in the reconstructed results. To mitigate this issue, we introduce a Thermal-Structural Guidance (TSG) module that explicitly encodes thermal semantics and structural cues as dual guidance. 

Based on the low-resolution input (\(\mathbf{I}_{\mathrm{LR}}\)), we construct two auxiliary representations: a heat map (\(\mathbf{I}_{\mathrm{Heat}}\)) to provide semantic heat-source information, and an edge map (\(\mathbf{I}_{\mathrm{Edge}}\)) to capture geometric boundaries, as shown in \cref{fig:pipeline}. Each map is processed through a dedicated encoder \(\mathbf{F}_{\mathrm{Heat}} = \mathrm{Enc}_{T}(\mathbf{I}_{\mathrm{Heat}})\) and \(\mathbf{F}_{\mathrm{Edge}} = \mathrm{Enc}_{S}(\mathbf{I}_{\mathrm{Edge}})\), where $\mathrm{Enc}_{T}$ and $\mathrm{Enc}_{S}$ are pre-trained encoders based on the DINOv3~\cite{simeoni2025dinov3}. To combine the two modalities, we use an adaptive weighting mechanism that fuses local and global cues. A learnable attention gate \(\mathbf{W} = \sigma(L(\mathbf{A}) + G(\mathbf{A}))\), where \(\mathbf{A} = \mathbf{F}_{\mathrm{Heat}} + \mathbf{F}_{\mathrm{Edge}}\), adaptively balances the contributions of thermal and structural features.
Here, \(L(\cdot)\) and \(G(\cdot)\) represent local and global attention operators, and \(\sigma(\cdot)\) denotes the sigmoid function. The fused guidance is computed as:
\begin{equation}
\mathbf{F}_{\mathrm{Fused}} \;=\; \,\mathbf{F}_{\mathrm{Heat}}\odot\mathbf{W}
\;+\; \,\mathbf{F}_{\mathrm{Edge}}\odot\!\left(\mathbf{1}-\mathbf{W}\right),
\label{eq:fused}
\end{equation}

\noindent where \(\odot\) denotes element-wise multiplication. This design enables spatially adaptive fusion, which regions with salient thermal patterns rely more on \(\mathbf{F}_{\mathrm{Heat}}\), and those with clear structural boundaries emphasize \(\mathbf{F}_{\mathrm{Edge}}\). 

The fused representation \(\mathbf{F}_{\mathrm{Fused}}\) serves as a semantic–structural layout prior to guide the low-resolution feature \(\mathbf{F}_{\mathrm{LR}} = \mathrm{Enc}_{I}(\mathbf{I}_{\mathrm{LR}})\). We employ a cross-attention module to propagate aligned information as \(\mathbf{F}_{\mathrm{TSG}} = \mathrm{Softmax}\!\left(\frac{QK^{\top}}{\sqrt{d}}\right)V\), where \(Q = \mathbf{W}_{Q}\mathbf{F}_{\mathrm{LR}}\), \(K =\mathbf{W}_{K}\mathbf{F}_{\mathrm{Fused}}\), and \(V = \mathbf{W}_{V}\mathbf{F}_{\mathrm{Fused}}\). \(\mathbf{W}_{Q}\), \(\mathbf{W}_{K}\), and \(\mathbf{W}_{V}\) are learnable linear projections, and \(d\) denotes the feature dimension for scaling. This prevents the model from overfitting to high-intensity thermal regions and encourages accurate boundary reconstruction. Empirically, this results in improved contour sharpness, reduced thermal drift, and enhanced physical interpretability in super-resolved infrared images.

\subsection{Condition-Adaptive Codebook}
Real infrared images often suffer from complex, spatially non-uniform degradations such as defocus blur, motion blur, and sensor noise~\cite{li2024contourlet, li2025difiisr}. These degradations not only distort local textures but also introduce code selection bias in autoregressive models, where incorrect discrete tokens may be activated under unstable degradation patterns. Moreover, the inherent quantization in VQ-VAE~\cite{van2017neural} introduces discretization errors that prevent the precise recovery of details, even with accurate code selection~\cite{kong2025nsarm, qu2025visual}. Consequently, the reconstructed images may exhibit over-smoothed textures and weakened structural fidelity.

To overcome these issues, we propose a Condition-Adaptive Codebook (CAC), which dynamically refines codebook embeddings based on degradation-aware priors. As shown in \cref{fig:pipeline}, instead of performing a static table lookup, each code embedding is adaptively modulated through low-rank perturbations conditioned on the low-resolution observation and cues such as thermal distributions and edge structures. Therefore, the same discrete index can decode to different embedding vectors under different degradation conditions and scenes. This design enables the model to adaptively refine decoded features while maintaining stable discrete semantics, thus reducing quantization bias and improving texture realism. Formally, the code embedding update process is defined as:
\begin{equation}
\mathbf{Z}'(g)[i] = \mathbf{Z}[i] + \tanh(\alpha)\,\big[(\mathbf{U}_i \odot \mathbf{h}(g))\mathbf{V}^{\top}\big],
\end{equation}

\noindent where \(\mathbf{Z}[i] \in \mathbb{R}^{d}\) denotes the basic embedding of the \(i\)-th code, \(\mathbf{U}_i \in \mathbb{R}^{r}\) is the low-rank basis vector for that code, \(\mathbf{V} \in \mathbb{R}^{d \times r}\) represents the shared feature direction matrix, and \(\mathbf{h}(g) \in \mathbb{R}^{r}\) is a condition vector derived from \(\mathbf{F}_{\mathrm{TSG}}\), incorporating cues such as thermal distributions and edge structures. \(\odot\) denotes element-wise multiplication, and \(\tanh(\alpha)\) is a gating factor that constrains the perturbation magnitude to ensure stable optimization.

By enabling code embeddings to vary adaptively with degradation conditions, the proposed CAC effectively mitigates code selection bias and quantization artifacts, yielding more structurally consistent and texture-rich infrared reconstructions under real-world degradations.

\subsection{Optimization}
Following the VAR~\cite{tian2024visual}, we employ a cross-entropy loss \(\mathcal{L}_{\mathrm{CE}}\) to supervise the Transformer-based autoregressive module at the token level. However, even with correct code selection, quantization in VQ-VAE~\cite{van2017neural} inevitably introduces discretization errors, leading to texture degradation and loss of fine details. To alleviate this issue, our Condition-Adaptive Codebook adaptively refines code embeddings, while an additional pixel-level reconstruction loss is introduced to provide continuous supervision. Specifically, an MSE loss \(\mathcal{L}_{\mathrm{MSE}}\) is applied between the SR and HR images to improve fidelity. Yet, real infrared degradations, such as defocus and motion blur, cause spatially varying peak shifts and local temperature compression, where simple pixel-wise MSE fails to restore physically accurate thermal distributions.

To address this, we propose a Thermal Order Consistency Loss \(\mathcal{L}_{\mathrm{TOC}}\) that enforces the monotonic relationship between temperature and pixel intensity. Unlike MSE, which relies on absolute values, \(\mathcal{L}_{\mathrm{TOC}}\) constrains relative thermal ordering between SR and HR pairs, penalizing cases where brightness order is reversed due to LR-HR misalignment and heat diffusion. This preserves local temperature gradients and stabilizes heat-source contrast. Formally, it is defined as:
\begin{equation}
\begin{split}
\mathcal{L}_{\mathrm{TOC}}
&= \frac{1}{|\Omega|} 
\sum_{(i,j) \in \Omega}
\mathrm{ReLU}\!\Big(
-\big[
(\mathbf{I}_{\mathrm{SR}}^{p}(i) - \mathbf{I}_{\mathrm{SR}}^{p}(j)) \\
&\hspace{4em}\times
(\mathbf{I}_{\mathrm{HR}}^{p}(i) - \mathbf{I}_{\mathrm{HR}}^{p}(j))
\big]
\Big),
\end{split}
\label{eq:toc}
\end{equation}

\noindent where \(\Omega\) denotes the set of adjacent patches, and \(\mathbf{I}_{\mathrm{SR}}^{p}\) and \(\mathbf{I}_{\mathrm{HR}}^{p}\) are the SR and HR patches, respectively. The ReLU term penalizes inverted thermal ordering between patch pairs, thereby enforcing physically consistent monotonicity of infrared intensity. This patch-wise formulation ensures robustness to minor spatial misalignments while maintaining correct local temperature ordering, effectively mitigating thermal peak drift in the reconstructed results. Finally, the overall training objective is defined as:
\begin{equation}
\mathcal{L}_{\mathrm{total}}
\;=\;
\mathcal{L}_{\mathrm{CE}}
+ \lambda_{1}\,\mathcal{L}_{\mathrm{MSE}}
+ \lambda_{2}\,\mathcal{L}_{\mathrm{TOC}},
\label{eq:total_loss}
\end{equation}

\noindent where \(\mathcal{L}_{\mathrm{CE}}\) supervises token prediction, \(\mathcal{L}_{\mathrm{MSE}}\) ensures pixel-level fidelity, and \(\mathcal{L}_{\mathrm{TOC}}\) enforces physical consistency of thermal distributions. The coefficients \(\lambda_{1}\) and \(\lambda_{2}\) balance the contribution of each term.

\begin{figure}[!tp]
    \centering
    \includegraphics[width=1\linewidth]{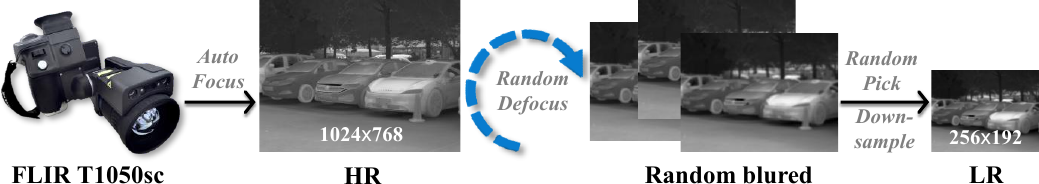}
    \caption{Data collection pipeline of FLIR-IISR.}
    \label{fig:dataset}
\end{figure}

\subsection{FLIR-IISR}
As shown in \cref{fig:teaser}, we construct FLIR-IISR, a real-world IISR dataset that contains \textbf{1,457} paired LR–HR images captured across \textbf{6} cities, \textbf{3} seasons, and \textbf{2} real blur patterns of optical and motion blur across \textbf{12} scene categories, stored in lossless BMP format to preserve radiometric fidelity. LR and HR pairs may exhibit slight sub-pixel misalignment due to the defocus–refocus acquisition process.

As shown in \cref{fig:dataset}, an automated capture program was developed to generate paired LR–HR data. Upon each electronic shutter trigger, the camera first performs automatic focusing to acquire a sharp HR image. Then, the program randomly adjusts the electronic focus ring to produce multiple defocus levels. One blurred frame is randomly selected and downsampled by 4× to obtain the LR counterpart. This process ensures consistent viewpoints while providing realistic degradations, including defocus and motion blur caused by moving objects.

FLIR-IISR provides dual-level annotations: degradation labels and scene labels. In the degradation dimension, the dataset contains 1,305 images with defocus blur and 152 with motion blur. In the semantic dimension, it covers 12 scene categories: person (309), bicycle (22), motorcycle (27), tricycle (13), car (234), bus (5), plane (54), statue (157), regular object (248), building (706), road (132), and complex scene (401), as shown in \cref{fig:teaser}. Note that an image may contain multiple scene categories. This hierarchical annotation makes FLIR-IISR a comprehensive and practical benchmark for evaluating real-world IISR performance.

%% file: sec/4_experiments.tex
\begin{table*}[!ht]
\centering
\small
\renewcommand{\arraystretch}{1.2}
\setlength{\tabcolsep}{1.2mm}
\begin{tabular}{lcc | cccc  cccc}
\Xhline{1.1pt}
\multicolumn{3}{c}{\textbf{Datasets}} & \multicolumn{2}{c}{\textbf{FLIR-IISR@Set5}} & \multicolumn{2}{c}{\textbf{FLIR-IISR@Set15}} & \multicolumn{2}{c}{\textbf{\(\text{M}^3\text{FD}\)@Set5}} & \multicolumn{2}{c}{\textbf{\(\text{M}^3\text{FD}\)@Set15}}\\ 
\cmidrule(lr){1-3} 
\cmidrule(lr){4-5} \cmidrule(lr){6-7} \cmidrule(lr){8-9} \cmidrule(lr){10-11}
 \multicolumn{3}{c}{Methods}  & MUSIQ$\uparrow$ & MANIQA$\uparrow$ & MUSIQ$\uparrow$ & MANIQA$\uparrow$ & MUSIQ$\uparrow$ & MANIQA$\uparrow$ & MUSIQ$\uparrow$ & MANIQA$\uparrow$\\
\cmidrule(lr){1-11} 
Low Resolution & - & -  & 27.9188 & 0.2030 & 25.4302 & 0.1740 & 24.2688 & 0.1438 & 24.7427 & 0.1770 \\
High Resolution  & -  & -  & 55.1375  & 0.3333  & 51.8250 & 0.3036 & 22.6813 & 0.1626 & 28.7760 & 0.2149  \\
\cmidrule(lr){1-11} 
HAT~\cite{chen2023activating} & CVPR'23 & ISR & 40.9844   & 0.2843  & 33.9990   & 0.2349  & 22.2281   & 0.1822   & 25.5260   & 0.2351   \\
BI-DiffSR~\cite{chen2024binarized} & NIPS'24 & ISR & 37.6250   & 0.2279   & 32.5854   & 0.1883   & 19.8031   & 0.1354  & 25.7417   & 0.1755   \\
PFT-SR~\cite{long2025progressive} & CVPR'25 & ISR & 41.1875   & 0.2847   & 33.9823   & 0.2366  & 22.3563   & 0.1769   & 25.7583   & 0.2393   \\
CoRPLE~\cite{li2024contourlet} & ECCV'24 & IISR & 32.3094   & 0.2256   & 27.8458  & 0.2101   & 23.6844  & 0.1802   & 25.9229  & 0.2426   \\
InfraFFN~\cite{qin2025infraffn} & KBS'25 & IISR & 36.8250   & 0.2393   & 30.9240   & 0.2226  & 22.1625   & 0.1766   & 25.0010   & 0.2346   \\
DifIISR~\cite{li2025difiisr} & CVPR'25 & IISR & \underline{54.7875} & 0.3672 & \underline{53.1625} & 0.3310 & 40.4563 & \textbf{0.2801} & 48.1625  & \underline{0.3248} \\
RealSR~\cite{ji2020real} & CVPR'20 & R-ISR & 41.1844   & 0.2883   & 39.5490   & 0.2543   & 21.1906   & 0.1572   & 27.6917  & 0.2261    \\
SinSR~\cite{wang2024sinsr} & CVPR'24 & R-ISR & 54.1625   & \underline{0.3719}   & 53.0854   & \underline{0.3342}   & \underline{40.9125}   & 0.2277  & \underline{48.3479}   & \textbf{0.3348}   \\
VARSR~\cite{qu2025visual} & ICML'25 & R-ISR & 52.7625   & 0.2948   & 51.9969   & 0.2995   & 38.9438   & \underline{0.2776}  & 39.9427   & 0.3003   \\
\rowcolor[gray]{0.92} Ours & - & R-IISR & \textbf{59.9000}   & \textbf{0.3776}   & \textbf{57.0625}   & \textbf{0.3403}   & \textbf{41.5750}   & 0.2532   & \textbf{49.0458}   & 0.3074   \\
\Xhline{1.1pt}
\end{tabular}
\caption{No-reference Metrics Comparison on FLIR-IISR and \(\text{M}^3\text{FD}\) datasets. The best is in \textbf{bold}, while the second is \underline{underlined}. For \(\text{M}^3\text{FD}\), Set5/15 are randomly sampled subsets. For FLIR-IISR, Set5/15 corresponds to motion/optical blur.}
\label{Table: compare_NRm}
\end{table*}

\begin{table*}[!ht]
\centering
\small
\renewcommand{\arraystretch}{1.2}
\setlength{\tabcolsep}{0.55mm}
\begin{tabular}{lcc | cccc  cccc cccc}
\Xhline{1.1pt}
\multicolumn{3}{c}{\textbf{Datasets}} & \multicolumn{3}{c}{\textbf{FLIR-IISR@Set5}} & \multicolumn{3}{c}{\textbf{FLIR-IISR@Set15}} & \multicolumn{3}{c}{\textbf{\(\text{M}^3\text{FD}\)@Set5}} & \multicolumn{3}{c}{\textbf{\(\text{M}^3\text{FD}\)@Set15}}\\ 
\cmidrule(lr){1-3} 
\cmidrule(lr){4-6} \cmidrule(lr){7-9} \cmidrule(lr){10-12} \cmidrule(lr){13-15}
 \multicolumn{3}{c}{Methods} & PSNR$\uparrow$ & SSIM$\uparrow$ & LPIPS$\downarrow$ & PSNR$\uparrow$ & SSIM$\uparrow$ & LPIPS$\downarrow$  & PSNR$\uparrow$ & SSIM$\uparrow$ & LPIPS$\downarrow$ & PSNR$\uparrow$ & SSIM$\uparrow$ & LPIPS$\downarrow$ \\
\cmidrule(lr){1-15} 
BI-DiffSR~\cite{chen2024binarized} & NIPS'24 & ISR & \underline{27.1976}  & 0.7869  & 0.4218   & \underline{28.7676}  & 0.8049   & 0.5206   & \textbf{35.9204}  & 0.9062  & 0.3235  & \textbf{34.8666 }& 0.8644  & 0.3133   \\
DifIISR~\cite{li2025difiisr} & CVPR'25 & IISR & 27.1969 & \underline{0.8195 }& 0.2525 & 28.5603 & 0.8474 & 0.2739 & \underline{35.8423 }& 0.9318 & 0.2474 & \underline{34.6620 }& \textbf{0.9114 }& 0.2214 \\
SinSR~\cite{wang2024sinsr} & CVPR'24 & R-ISR & 26.7594   & 0.6970   & 0.3670   & 28.3163   & 0.8521   & 0.2956   & 35.1806  & \underline{0.9323}  & 0.2528  & 34.0048  & 0.9041 & \underline{0.1652}   \\
VARSR~\cite{qu2025visual} & ICML'25 & R-ISR & 26.9767   & 0.7868   & \underline{0.2304}   & 28.3439   & \underline{0.8613}   & \underline{0.2003}   & 33.6762  & 0.9268  & \underline{0.2436}  & 32.6001  & 0.8997  & 0.1895   \\
\rowcolor[gray]{0.92} Ours & - & R-IISR & \textbf{28.5126}  & \textbf{0.8278}  & \textbf{0.1615}   & \textbf{29.5136}  & \textbf{0.8895}  & \textbf{0.1340} 
  & 32.3175  & \textbf{0.9383}  & \textbf{0.1997}  & 31.5633  & \underline{0.9047 } & \textbf{0.1361}   \\
\Xhline{1.1pt}
\end{tabular}
\caption{Reference-based Metrics Comparison on FLIR-IISR and \(\text{M}^3\text{FD}\) datasets. The best is in \textbf{bold}, while the second is \underline{underlined}. For \(\text{M}^3\text{FD}\), Set5/15 are randomly sampled subsets. For FLIR-IISR, Set5/15 corresponds to motion/optical blur.}
\label{Table: compare_Rm3}
\end{table*}

\section{Experiments}
\subsection{Experimental Settings}
\noindent\textbf{Implementation Details.} 
Our Real-IISR was trained on 4 NVIDIA A800 GPUs. To accelerate the training, we utilized the pre-trained VAR and VQVAE from~\cite{qu2025visual}. The training process employs the AdamW~\cite{loshchilov2017fixing} optimizer with a batch size of 4, a weight decay of $5\times10^{-2}$, and a learning rate of $5\times10^{-5}$. The model is fine-tuned for 10k iterations. For loss balancing, we set \(\lambda_{1} = 0.2\) and \(\lambda_{2} = 0.8\). In computing the Thermal Order Consistency Loss \(\mathcal{L}_{\mathrm{TOC}}\), the patch size is empirically set to 8. For the Condition-Adaptive Codebook, the rank of the low-rank modulation is set to $r=8$ in all experiments.

\noindent\textbf{Datasets and Metrics.} 
We train our model on the FLIR-IISR dataset with 1,192 images for training and 265 for testing, and evaluate it together with 500 test images selected from the M$^3$FD~\cite{liu2022target} dataset. Following the experimental settings in~\cite{qu2025visual, wang2024exploiting}, the resolution of HR images is set to 512$\times$512, while that of LR images is set to 128$\times$128 during both training and testing. To mitigate overfitting, random cropping is employed as a data augmentation strategy. 

For a comprehensive quantitative evaluation, we adopt three reference-based and two no-reference image quality metrics. To assess the reconstruction fidelity with respect to HR images, we compute PSNR, SSIM~\cite{wang2004image}, and LPIPS~\cite{zhang2018unreasonable}. In addition, to evaluate the perceptual quality of the generated images without reference, we employ MUSIQ~\cite{ke2021musiq} and MANIQA~\cite{yang2022maniqa}.

\noindent\textbf{Comparitive Methods.}
We conduct a comprehensive comparison with nine state-of-the-art methods, including three image super-resolution (ISR) methods: HAT~\cite{chen2023activating}, BI-DiffSR~\cite{chen2024binarized}, and PFT-SR~\cite{long2025progressive}; three infrared image super-resolution (IISR) methods: CoRPLE~\cite{li2024contourlet}, InfraFFN~\cite{qin2025infraffn}, and DifIISR~\cite{li2025difiisr}; and three real-world image super-resolution (R-ISR) methods: RealSR~\cite{ji2020real}, SinSR~\cite{wang2024sinsr}, and VARSR~\cite{qu2025visual}. All comparative methods are retrained on the FLIR-IISR dataset with the same setting as Real-IISR.

\subsection{Quantitative Comparison}
\noindent \textbf{No-reference Metrics Comparison.} As shown in \cref{Table: compare_NRm}, our method is compared with nine competitive approaches on the FLIR-IISR and $\text{M}^3\text{FD}$ datasets. It achieves the highest scores on both metrics for the Set5 and Set15 subsets of FLIR-IISR, attains the best MUSIQ performance on $\text{M}^3\text{FD}$, and maintains competitive results on MANIQA. The improvement in MUSIQ demonstrates that Real-IISR possesses a stronger capability in modeling global perceptual quality and structural consistency. By integrating thermal order consistency with structural priors during autoregressive reconstruction, Real-IISR effectively mitigates the coupled optical and perceptual degradations inherent in infrared imaging. Meanwhile, the stable MANIQA performance indicates that the model successfully preserves high-quality textures and edge details.

\noindent \textbf{Reference-based Metrics Comparison.} \cref{Table: compare_Rm3} presents quantitative comparisons on the FLIR-IISR and $\text{M}^3\text{FD}$ datasets using reference-based metrics. Real-IISR achieves the best overall performance across all subsets, demonstrating strong pixel-wise fidelity and perceptual consistency. This advantage verifies the effectiveness of the proposed Thermal Order Consistency Loss \(\mathcal{L}_{\mathrm{TOC}}\) that enforces the monotonic relationship between temperature and pixel intensity, and the Condition-Adaptive Codebook, which enhances texture realism under spatially variant degradations. Benefiting from the joint modeling of thermal radiation and structural information, Real-IISR consistently reconstructs geometrically clear images under complex real-world degradations, achieving a well-balanced trade-off between structural fidelity and perceptual consistency.

\begin{figure}[!ht]
 \centering
 \includegraphics[width=0.47\textwidth]{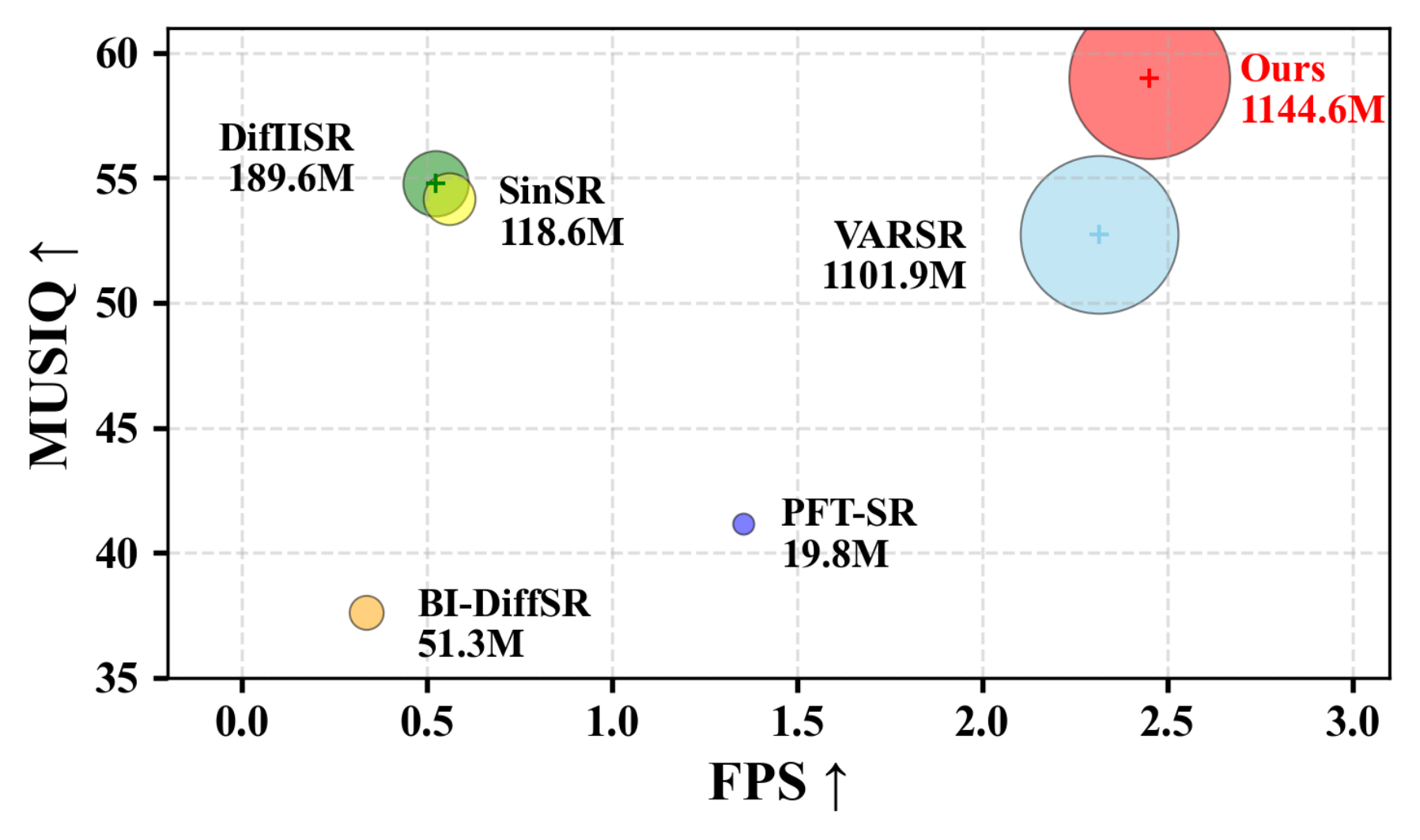}
    \caption{Efficiency comparison in terms of perceptual MUSIQ and FPS; circle diameter indicates model parameters.}
    \label{fig:E1}
\end{figure}

\begin{figure*}[!ht]
 \centering
    \includegraphics[width=1\textwidth]{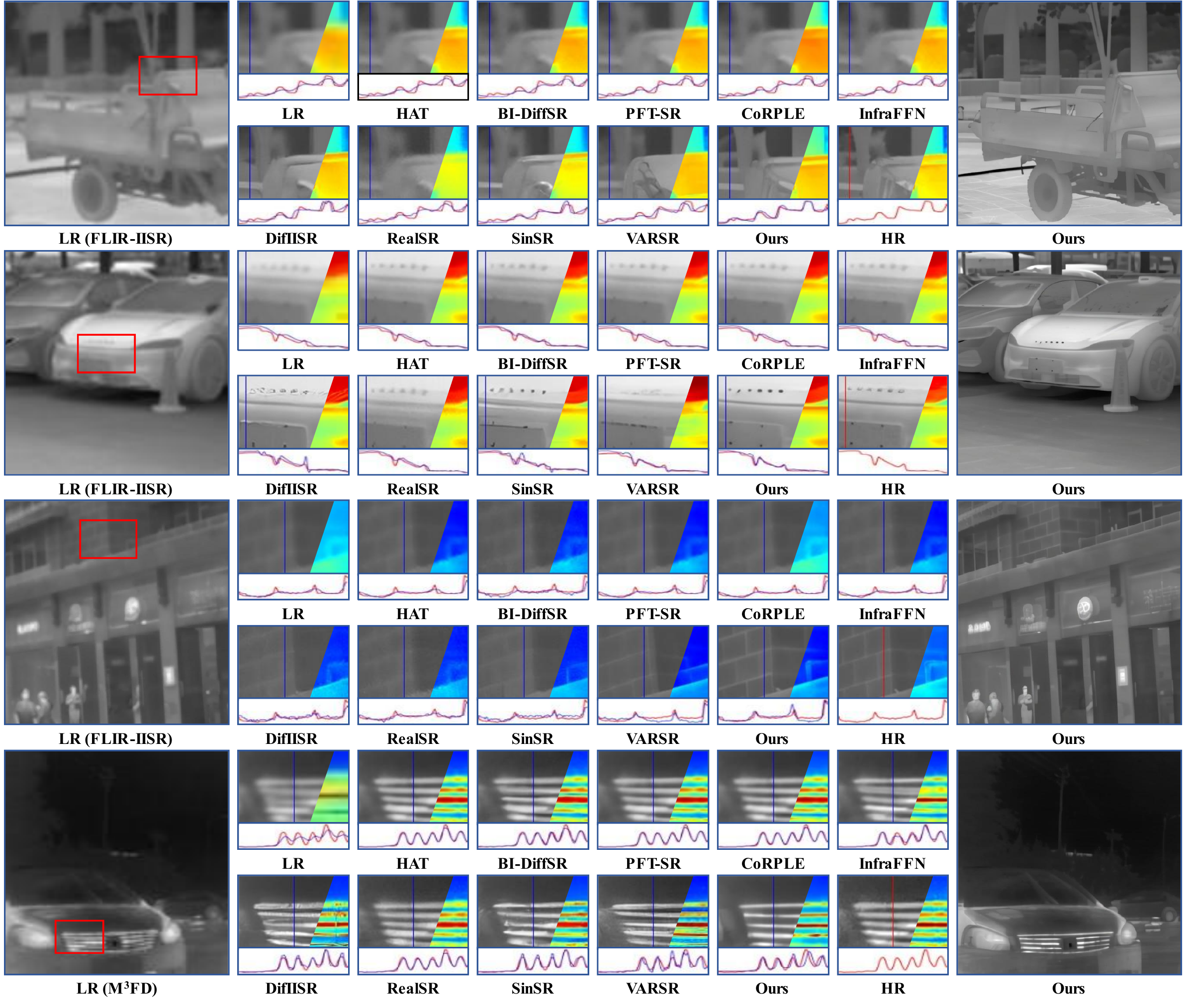}
    \caption{Qualitative comparison of IISR with SOTA methods on FLIR-IISR and \(\text{M}^3\text{FD}\) datasets. The graph illustrates grayscale fluctuations along the \textcolor{blue}{blue}-marked sampling line, and \textcolor{red}{red}-marked sampling line denotes the HR.}
    \label{fig:compare_31}
\end{figure*}

\subsection{Qualitative Results}
\cref{fig:compare_31} shows the qualitative comparisons on FLIR-IISR and $\text{M}^3\text{FD}$ datasets. For intuitive visualization, pseudo-color thermal maps are displayed to highlight the radiometric consistency of different methods, and the graph illustrates grayscale fluctuations along the sampling line. Competing methods often produce blurred contours, unstable thermal regions, or over-enhanced hotspots due to insufficient thermal perception. In contrast, Real-IISR reconstructs sharper edges and faithful heat distributions with fewer artifacts, and the Thermal Order Consistency Loss \(\mathcal{L}_{\mathrm{TOC}}\) further ensures the monotonic relationship between temperature and pixel intensity to avoid thermal peak drift. For instance, in the water pipe region of the third row, IISR methods (e.g., DifIISR and CoRPLE) tend to generate unstable thermal boundaries, while R-ISR methods (e.g., VARSR and RealSR) fail to preserve the correct temperature and generate notable artifacts. The Thermal-Structural Guidance aligns heat radiation with object boundaries, while the Condition-Adaptive Codebook enhances fine-grained texture details under spatially varying degradations. Real-IISR preserves both edge sharpness and thermal uniformity, demonstrating its advantage in maintaining structural fidelity and physically consistent thermal patterns.

\subsection{Efficiency Analysis}
We evaluate model efficiency in terms of parameters and inference speed (FPS) on a single NVIDIA A800 GPU. As shown in \cref{fig:E1}, Real-IISR, though the largest model (1144.6 M), achieves the fastest inference (2.45 FPS) and best perceptual quality. Diffusion-based methods are slowed by multi-step denoising, while autoregressive frameworks enable deterministic generation with higher throughput. Compared to the VAR-based VARSR~\cite{qu2025visual} that adopts a diffusion-based refiner to gradually refine the generated results, Real-IISR achieves \textbf{6\% faster} inference despite being slightly larger in size, owing to its concise architecture, demonstrating superior computational efficiency.

\subsection{Ablation Study}

\begin{figure}[!t]
 \centering
 \includegraphics[width=0.47\textwidth]{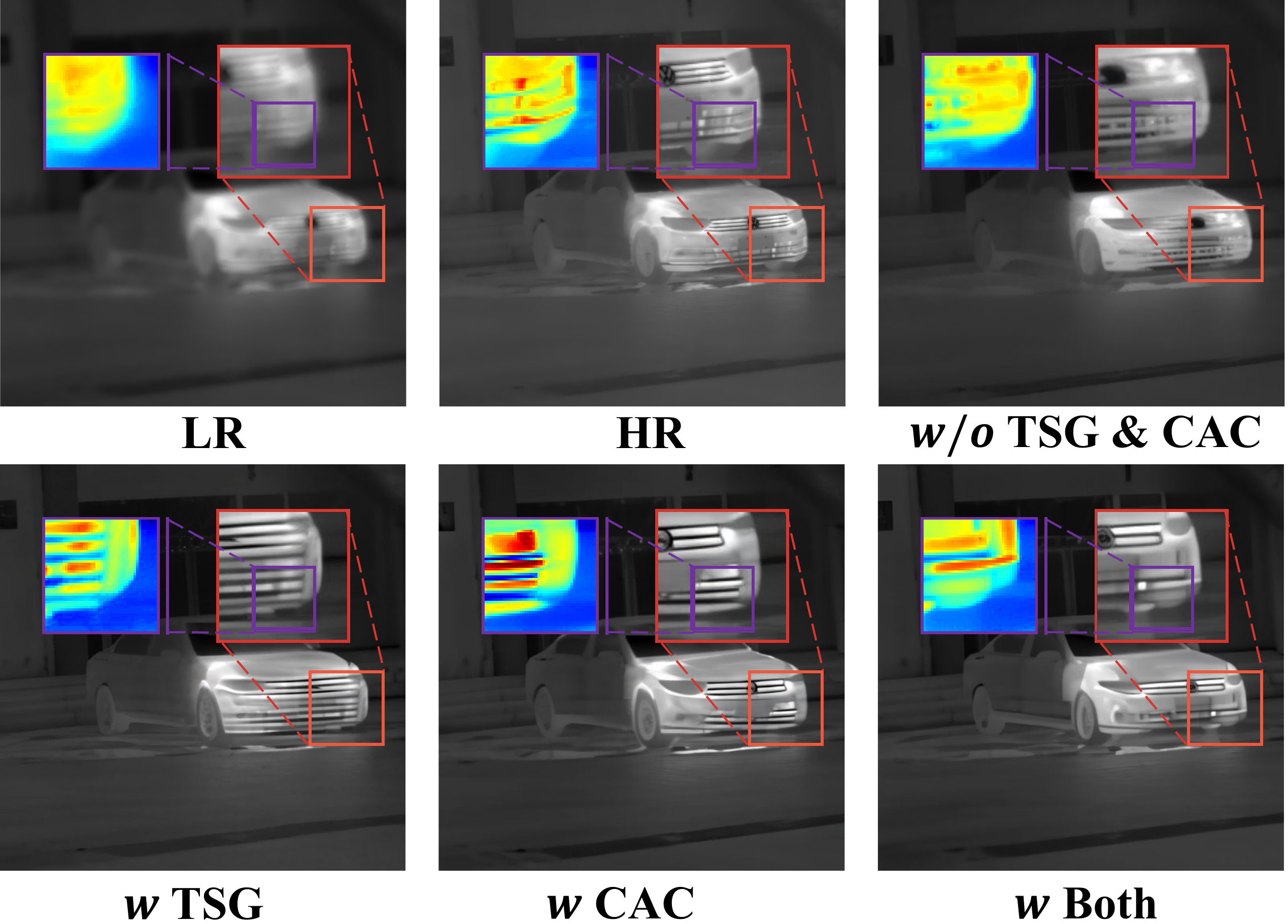}
    \caption{Qualitative ablation on the Thermal-Structural Guidance (TSG) and Condition-Adaptive Codebook (CAC).}
    \label{fig:A1}
\end{figure}

\begin{figure}[!t]
 \centering
 \includegraphics[width=0.47\textwidth]{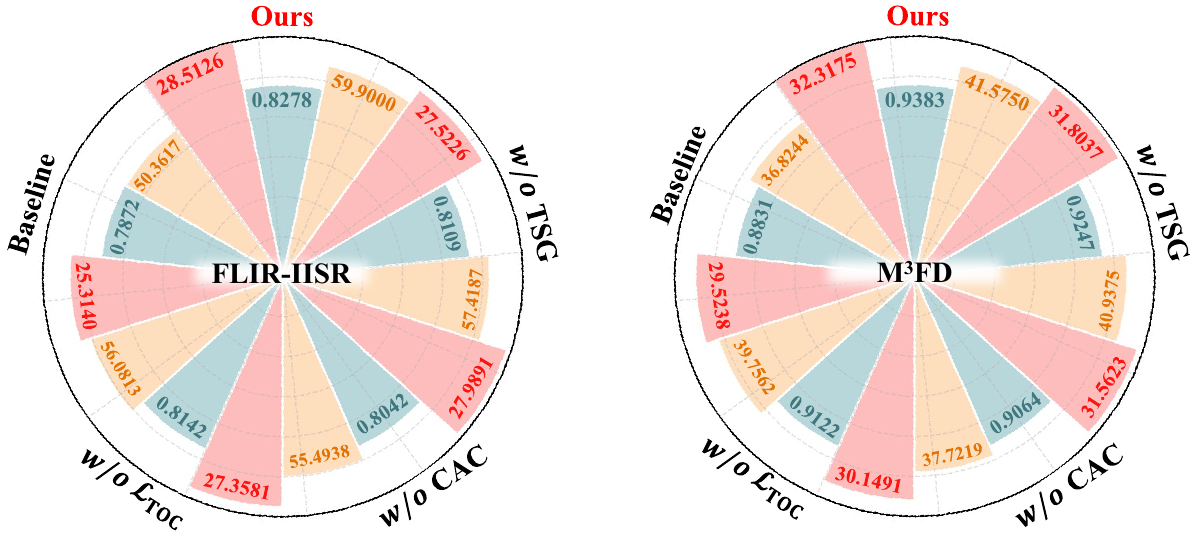}
    \caption{Quantitative ablation of TSG, CAC, and \(\mathcal{L}_{\text{TOC}}\) on \textcolor{red}{PSNR}, \textcolor{SkyBlue}{SSIM}, and \textcolor{Goldenrod}{MUSIQ}.}
    \label{fig:abla_tsg_cac}
\end{figure}

\noindent\textbf{Effectiveness of TSG and CAC.}
To validate the contribution of the proposed Thermal-Structural Guidance (TSG) and Condition-Adaptive Codebook (CAC), we perform an ablation study on the FLIR-IISR and \(\text{M}^3\text{FD}\) datasets, as summarized in \cref{fig:abla_tsg_cac} and visualized in \cref{fig:A1}. Removing TSG leads to inaccurate alignment between thermal radiation and object boundaries, causing blurred edges and weakened structural contours. Excluding CAC results in unstable textures and inconsistent heat distributions, reflected by a notable degradation in MUSIQ and SSIM. In contrast, combining both modules achieves the best overall performance. As shown in \cref{fig:A1}, our full model reconstructs sharper boundaries and faithful temperature, highlighting its effectiveness in maintaining structural fidelity and radiometric consistency.

\begin{figure}[!t]
 \centering
 \includegraphics[width=0.47\textwidth]{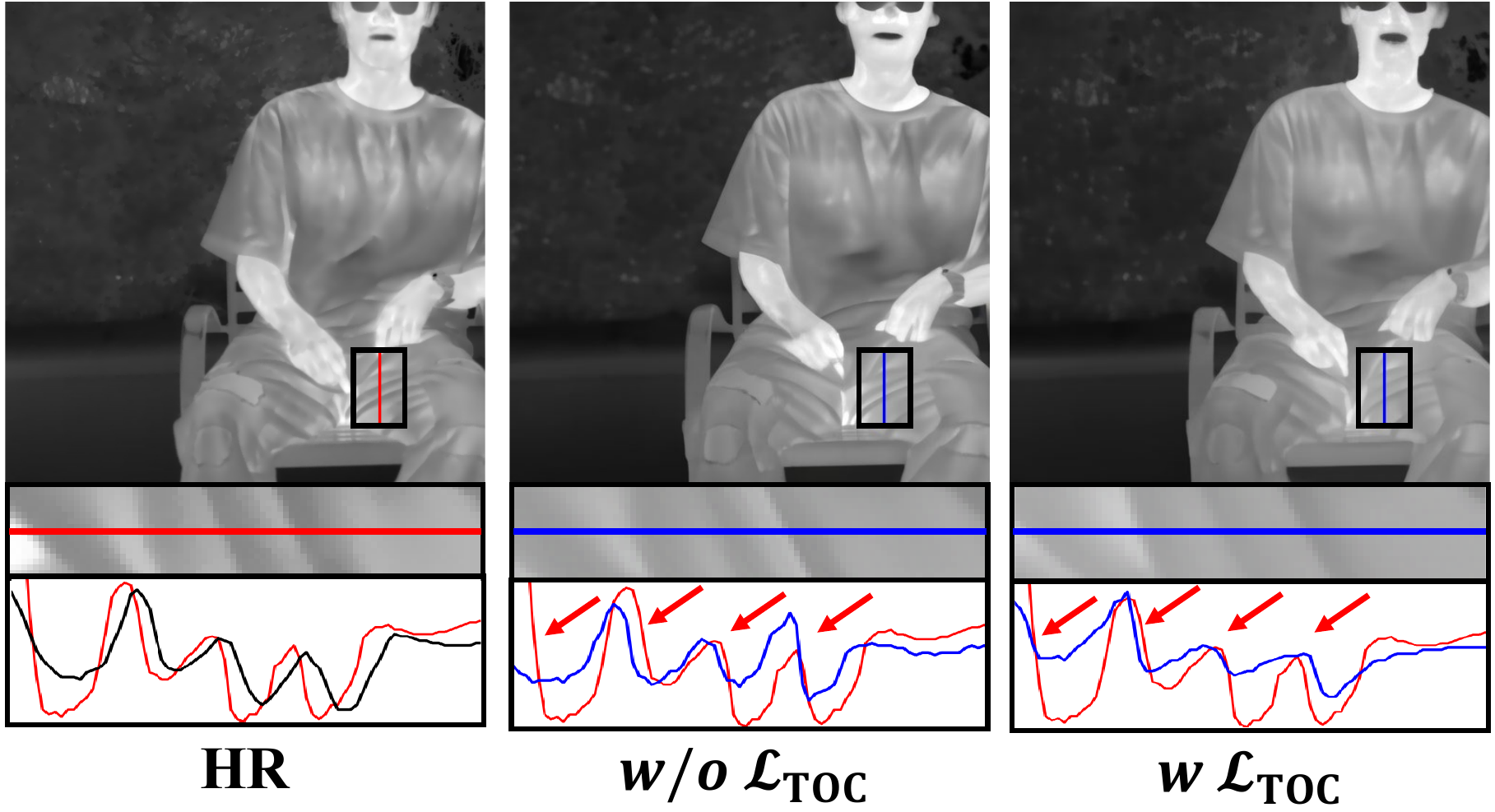}
    \caption{Qualitative ablation on the Thermal Order Consistency Loss \(\mathcal{L}_{\text{TOC}}\) with graph illustrates grayscale fluctuations along \textcolor{red}{HR}, \textcolor{black}{LR}, and \textcolor{blue}{models}.}
    \label{fig:A2}
\end{figure}

\noindent\textbf{Impact of Loss.} To assess the effectiveness of the Thermal Order Consistency Loss \(\mathcal{L}_{\text{TOC}}\), we conduct comparative experiments with and without this constraint. As shown in \cref{fig:A2}, removing the \(\mathcal{L}_{\text{TOC}}\) leads to disrupted thermal intensity ordering, resulting in thermal peak drift and local temperature 
compression. Correspondingly, \cref{fig:abla_tsg_cac} shows a notable drop in MUSIQ scores. Incorporating \(\mathcal{L}_{\text{TOC}}\) preserves the monotonic brightness–temperature relationship, yielding smoother and more coherent thermal distributions with improved physical consistency. This constraint effectively mitigates thermal peak shifts caused by non-uniform degradations, ensuring the radiometric reliability and visual stability of the reconstructed infrared images.

\begin{figure}[!t]
 \centering
 \includegraphics[width=0.47\textwidth]{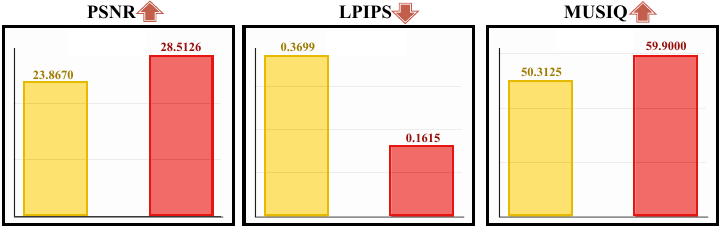}
    \caption{Quantitative ablation on the baseline choice of \textcolor{Goldenrod}{diffusion} and \textcolor{red}{VAR}.}
    \label{fig:A3}
\end{figure}

\noindent\textbf{Choice of Generation Baseline.} To examine the influence of the underlying generative paradigm, we replace the VAR backbone with a diffusion-based architecture (ResShift~\cite{yue2023resshif}) while keeping the same Thermal-Structural Guidance (TSG) as conditional input. As shown in \cref{fig:A3}, the diffusion-based variant fails to produce accurate results for both no-reference and reference-based evaluation, whereas the VAR-based framework achieves higher PSNR (28.51), lower LPIPS (0.1615), and better MUSIQ (59.90). This performance gap arises because iterative denoising in diffusion models tends to blur high-frequency thermal details and misalign structural cues, while the deterministic, token-level prediction in VAR preserves fine textures and consistent thermal–structural correspondence. These results verify that autoregressive generation better matches the discrete and spatially structured nature of real-world infrared imaging.

%% file: sec/5_conclusion.tex
\section{Conclusion}
We tackle the challenge of real-world IISR, where synthetic degradations limit generalization to real scenarios. To bridge this gap, we introduce \textbf{FLIR-IISR}, a real-world paired dataset covering diverse scenes and physical degradations. Building on it, we propose \textbf{Real-IISR}, a unified autoregressive framework that aligns thermal and structural cues, adapts a Condition-Adaptive Codebook, and enforces the thermal ordering. Extensive experiments on both real and synthesised datasets show the impressive performance of our Real-IISR.

\noindent \textbf{Broader Impact.} 
The proposed FLIR-IISR dataset offers a new real-world benchmark for investigating infrared imaging degradations, thereby promoting progress toward realistic infrared restoration. Building on it, Real-IISR enhances thermal perception and structural fidelity, benefiting applications in autonomous driving, surveillance, and thermal monitoring under adverse conditions, establishing a solid foundation for realistic infrared restoration.